%% file: main.tex
\documentclass[letterpaper, 10 pt, conference]{ieeeconf}  
\IEEEoverridecommandlockouts                              
\usepackage{amsmath} 
\usepackage{graphicx}
\usepackage{ulem}
\usepackage{changepage}
\usepackage[mathletters]{ucs}
\usepackage[utf8x]{inputenc}
\usepackage{bm}
\usepackage{xcolor}
\usepackage{cite}
\usepackage{float}
\usepackage{color}
\usepackage{balance}
\usepackage{tabularx} 

\title{\LARGE \bf
DDPEN: Trajectory Optimisation With Sub Goal Generation Model
}
\author{Aleksander Gamayunov$^{1,2}$ \and Aleksey Postnikov$^{1,2}$ \and Gonzalo Ferrer$^{2}$%
        \thanks{\textsuperscript{1} The authors are with the Sberbank Robotics Laboratory, Moscow, Russia.
                    {\tt\small \{gamayunov.a.r, postnikov.a.l\}@sberbank.ru}.
               }
        \thanks{\textsuperscript{2}Skolkovo Institute of Science and Technology, Moscow, Russia.
                    {\tt\small g.ferrer@skoltech.ru}.
           }    
    }

\begin{document}

\maketitle
\thispagestyle{empty}
\pagestyle{empty}

\input{chapters/0_abstract}
\input{chapters/1_intro}
\input{chapters/3_method}

\input{chapters/4_experiments}

\input{chapters/5_conclusion}
\bibliographystyle{IEEEtran}
\bibliography{main}

\end{document}

%% file: chapters/0_abstract.tex
\begin{abstract}
    Differential dynamic programming (DDP) is a widely used and powerful trajectory optimization technique,  however, due to its internal structure, it is not exempt from local minima.
    In this paper, we present Differential Dynamic Programming with Escape Network (DDPEN) - a novel approach to avoid DDP local minima by utilising an additional term used in the optimization criteria pointing towards the direction where robot should move in order to escape local minima. 
    \par
    In order to produce the aforementioned directions, we propose to utilize a deep model that takes as an input the map of the environment in the form of a costmap together with the desired goal position.
    The Model produces possible future directions that will lead to the goal, avoiding local minima which is possible to run in real time conditions.
    The model is trained on a synthetic dataset and overall the system is evaluated at the Gazebo simulator.
    \par
    In this work we show that our proposed method allows avoiding local minima of trajectory optimization algorithm and successfully execute a trajectory 278 m long with various convex and nonconvex obstacles.
    
    
\end{abstract}


%% file: chapters/1_intro.tex
\section{Introduction}
\par
The need for service professions - couriers, cleaners, etc. is growing along with the population of the planet.
At the same time, mobile robotics continue to develop and can replace at least part of the monotonous work, while providing people with new highly skilled jobs.

We are witnessing news about novel robot couriers, and robot vacuum cleaners that have already become commonplace while autonomous tractors are starting to drive through the agricultural fields.

Navigation is one of the core subsystems for robot's autonomy which is largely dependent on path planning and trajectory planning algorithms.


In this work, we consider optimizing algorithms and propose a solution to the important but common problem of nonconvex trajectory optimization.

\par
Optimization trajectory algorithms iteratively modify the initial assumption of the trajectory according to the specified criteria, and gradually reduce the error value. 
One of the important issues is that the free state space is usually nonconvex and presents multiple local minima. 
An optimization algorithm can reach one of the local minima and stay in it without ever reaching the global minima.
\par
For a trajectory optimization task, reaching local minima can lead to a situation where the algorithm will not lead the mobile robot to the goal, even if the optimal trajectory exists.
\begin{figure}[tp]
    \centering
    \includegraphics[width=8.5cm]{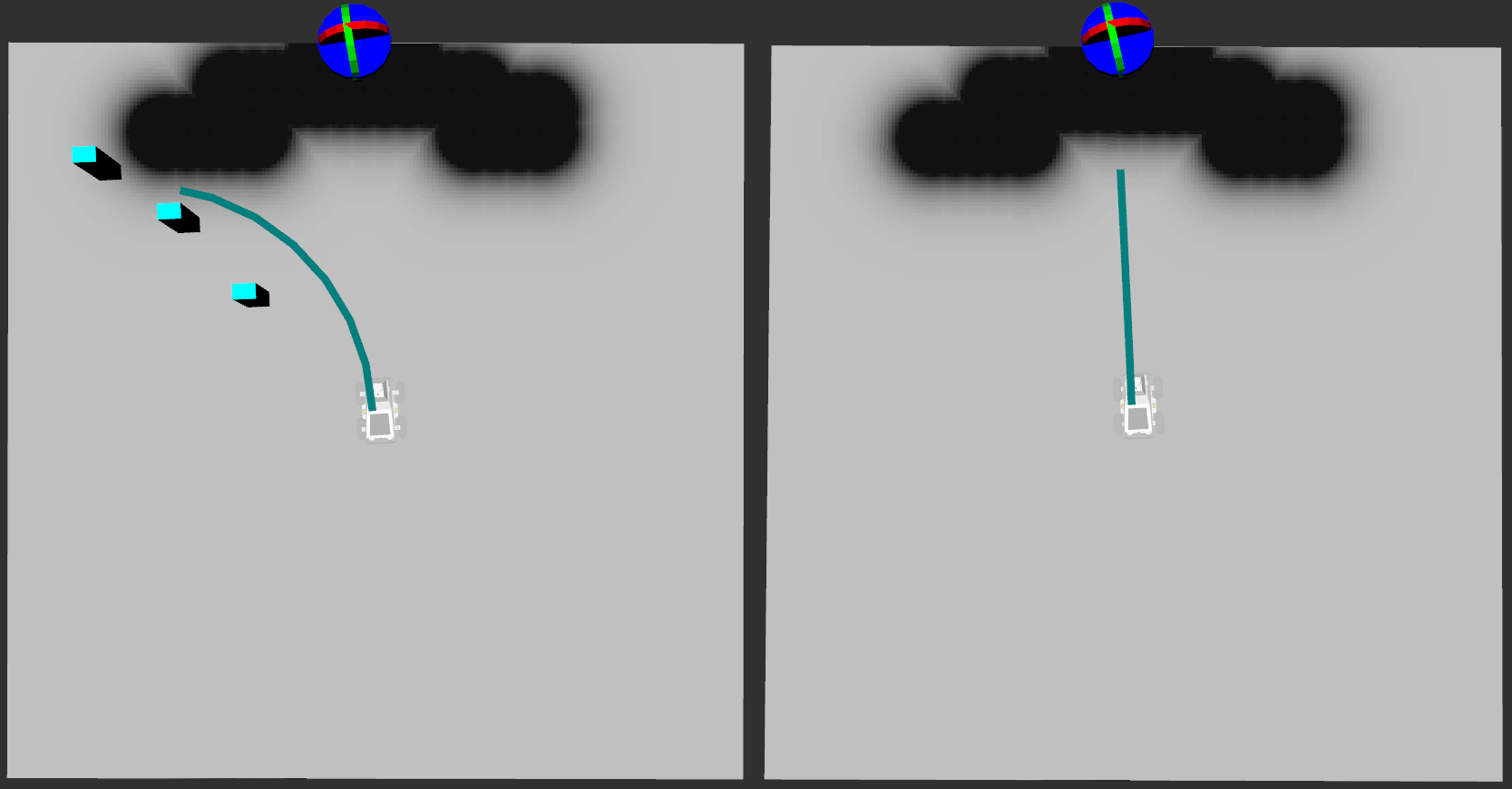}
    \caption{Comparison of DDPEN (left side) with DDP (right side) on trajectory optimization with nonconvex obstacle. The robot is in the center of gray costmap. Black part of the costmap represents obstacles. Green line is the optimized trajectory. Green rectangles is predicted sub goals $x_{s}$.  Blue circle is desired goal position $x_{goal}$.}
    \label{fig::DDP_vs_DDPEN}
\end{figure}
In that case, the mobile robot will not be able to complete its task. 

A number of works are devoted to this problem. Zhaoming Xie et al. in his work of extension the differential dynamic programming trajectory optimization approach (DDP\cite{DDP_origin_book_article,DDP_origin_book}) with nonlinear constraints (CDDP)\cite{CDDP}, mention that ``like all nonconvex optimization, we need a good initial trajectory to ensure that the algorithm would not get stuck in a bad local minima''.

In that case sampling trajectory planning algorithm like $RRT^*$\cite{RRT-star-quick} or some sort of fast neural path planning $A^*$\cite{neural_Astar} algorithms for generating a safe initial trajectory for optimization can be used.
Another approach of local minima avoidance is based on DDP maximum entropy formulation (MEDDP)\cite{MEDDP}, with multimodal policy exploration. 
\par
Escaping from local minima is a trivial task for human in case of 2d environment but the overall process of escaping from the local minimum is ill-posed with the current gradient-based optimisation tools in DDP. 
\par
In this work, we have proposed an approach for escaping from local minima that provides additional criteria of optimisation, calculated by a neural network, that ``suggests'' the best direction for the optimization process.

The Neural network learns to imitate optimal planning algorithm and provides intermediate goal as additional optimisation criteria that leads the optimisation process out from local minima. 

We have evaluated vanilla DDP optimisation algorithm and our proposed approach (DDPEN) that takes additional optimization criteria from neural network at simulated scenarios in Gazebo\cite{Gazebo} and have shown that our proposed approach is being able to complete complex trajectories even with nonconvex obstacles. 
\par

%% file: chapters/3_method.tex
\newcommand\norm[1]{\left\lVert#1\right\rVert}
\section{Method}
\subsection{Trajectory optimisation problem}
The state of an agent at time $t$ is represented by $x_t$, where $x$ is the two dimensional vector in the Cartesian coordinate system. 
Transitions between states are provided by the dynamic function $x_{t+1}=f(x_t,u_t)$ where $u_t$ is a control action at time $t$. 
Sequence of control action   $U=[u_{t_0},u_{t_1},...,u_{t_{N-1}}]$ applied to dynamic function forms the trajectory $X=[x_{t_0},x_{t_1},...,x_{t_N}]$ from initial time $t_0$ to prediction horizon $t_{N}$. 
The total cost of trajectory $J_0$ is the sum of running costs $l$ and the final cost $l_f$. The solution of the optimal control problem is to find the minimum cost control sequence.

\begin{equation}
    J_0(X,U)=\sum_{i=0}^{N-1}l(x_i,u_i)+l_f(x_N)
    \label{eq::total_cost}
\end{equation}
\begin{equation}
    U^*(X)\equiv \textit{arg}\min_{_{U}} J_0(X,U)
    \label{eq::optimal_control}
\end{equation}
\par
The current work focuses on the cost function $l(x,u)$ that forms the optimisation direction and can move the optimisation process into a local minimum. 

The cost function, in our vanilla realisation of DDP optimisation method, includes $c_g$ - cost of distance to the goal state $x_{goal}$, $c_c$ - control cost and $c_o$ - cost representing the distance to obstacles $obs$ at distance less than $d$ meters.
\begin{equation}
    l(x,u)=c_{c}(u) + c_{g}(x,x_{goal}) + c_{o}(x,obs)
    \label{eq::running_cost}
\end{equation}
\begin{equation}
    l_f(x)=c_{g}(x,x_{goal}) + c_{o}(x,obs)
    \label{eq::running_cost}
\end{equation}

\begin{equation}
    c_{c}(u)=\sum u^2
    \label{eq::cost_controll}
\end{equation}
\begin{equation}
    c_{g}(x,x_{goal})=\norm{x-x_{goal}}_2
    \label{eq::cost_goal}
\end{equation}
\begin{equation}
    c_{o}(x,obs,d)=\sum ReLU(d - \norm{x-obs}_2)
    \label{eq::cost_obstacles}
\end{equation}
\par
\par

\subsection{DDPEN}

The proposed method DDPEN takes into account an additional cost $c_{s}$ - distance to sub goal $x_{s}$, generated by the neural network Eq \ref{eq::x_intergoal}, which represents the movement direction of the mobile robot that allows avoiding local minima.
\begin{equation}
    l(x,u)=c_{c}(u) + c_{g}(x,x_{goal}) + c_{o}(x,obs) + c_{s}(x,x_{s})
    \label{eq::DDPEN_running_cost}
\end{equation}
\begin{equation}
    c_{s}(x,x_{s}) = \norm{x-x_{s}}_2
\label{eq::cost_intergoal}
\end{equation}

\begin{figure}[tp!]
    \includegraphics[width=8.5cm]{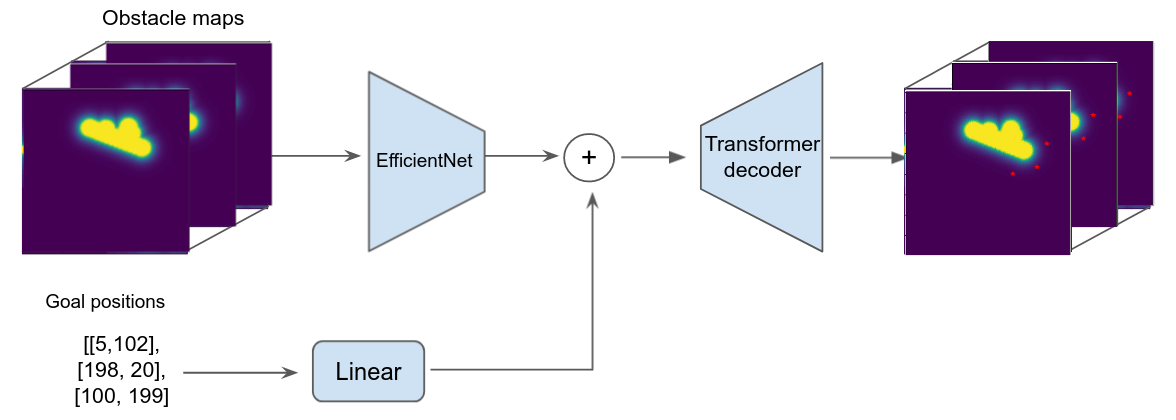}
    \caption{Sub Goal Generation Model architecture.}
    \label{fig::model}
\end{figure}

\subsection{Sub Goal Generation Model}

In order to generate future positions that will force the trajectory optimization algorithm to avoid local minima we propose to utilise deep model Eq. \ref{eq::x_intergoal}, that is trained to imitate goal proposals generated by $A^*$ algorithm, that would satisfy real-time restrictions and whose performance would not depend on the position of obstacles.

\begin{equation}
    x_{s} = SG_\theta(map,x_{goal}),
    \label{eq::x_intergoal}
\end{equation}
where $SG_\theta$ - deep  Sub Goal Generation Model, \textit{map} - local costmap of environment, $x_{goal}$ - desired goal position of robot.
\par
Architecture of Sub Goal Generation Model, shown in Figure \ref{fig::model}, consist of convolution based encoder of obstacle map, namely EfficientNet B0\cite{tan2019efficientnet}, linear layer that embeds goal positions, and transformer based decoder that generates sub goal positions.

\subsection{Dataset}

We have created a synthetic dataset that captures randomly generated obstacle maps and optimal path to randomly sampled goal on that map.
As a base for obstacle map generation it is taken a costmap with 200x200 cells with the cell side length of 5cm.
Predefined obstacles are created at random positions with non-blocking center position constraint.
Obstacles are positioned through a random affine transformation for augmentation purposes. 
After generation of obstacles, their costs are dilated from occupied cost to free moving space.
\par

\begin{figure}[tp]
    \centering
    \includegraphics[width=7.0cm]{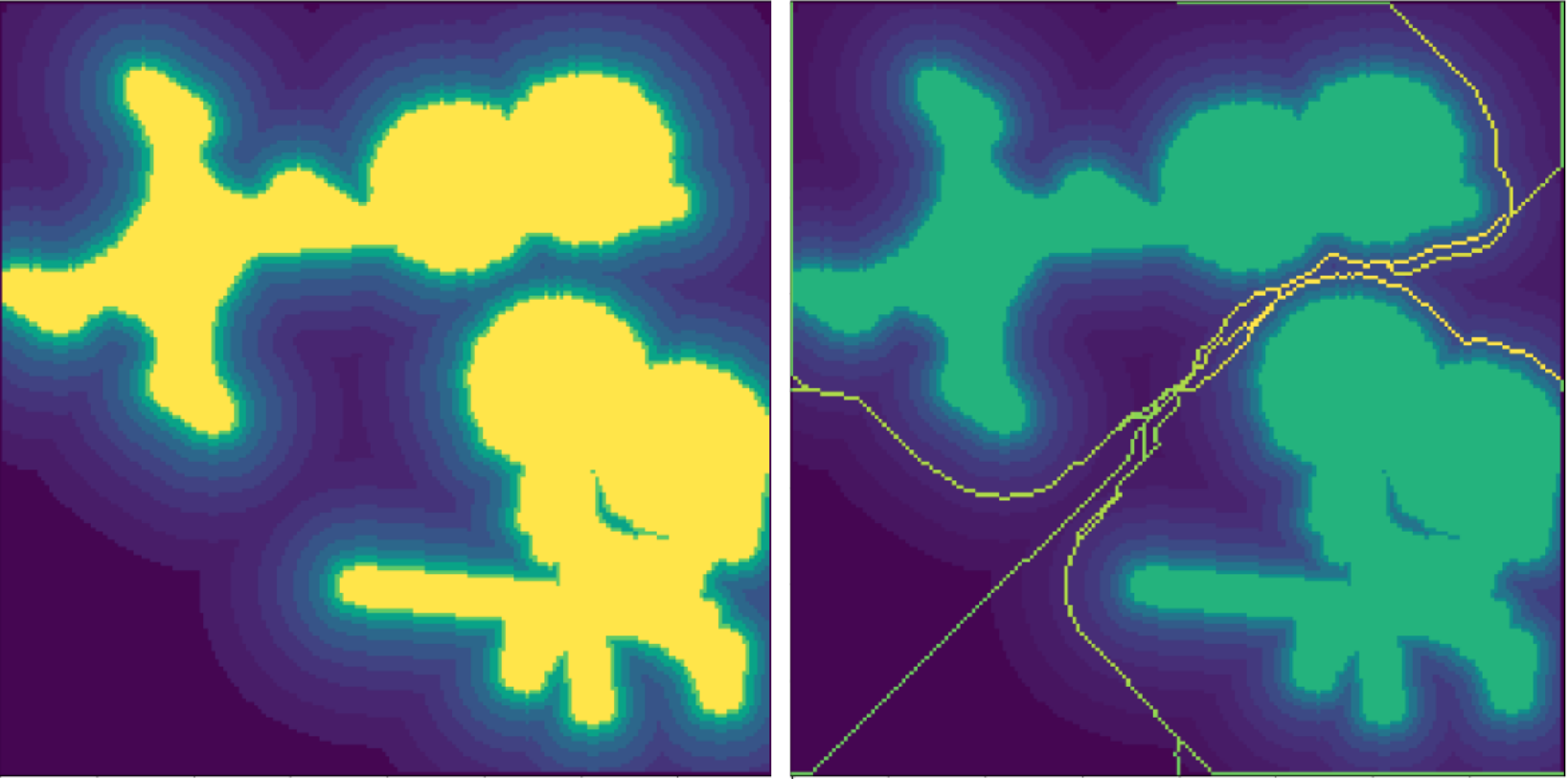}
    \caption{Synthetically generated samples from dataset. \textit{Left}: Example of randomly generated costmap   \textit{Right}: Example of paths obtained by A* algorithm to randomly sampled goals.}
    \label{fig::costmap_with_A_star}
\end{figure}

We have used $A^{*}$ path planning algorithm to generate examples of paths from starting position, which is in our case always the grid center, to randomly sampled goal position near borders of the costmap.
After processing the optimal path from randomly sampled goal positions on a randomly generated costmap, intermediate positions of sub goals at 30, 50, 70 steps of the optimal path are saved together with goals and costmaps, forming the overall dataset.
Synthetically generated samples from the dataset is shown in Figure \ref{fig::costmap_with_A_star}.


\begin{table*}[ht]
    \caption{Comparison of DDP and DDPEN time execution of path. The values is the mean$\pm\sigma$ in  seconds}
    \label{table:results}
    \begin{center}
        \begin{tabular}{|| c | c c| c c|} 
         \hline
         Type of obstacles & \multicolumn{2}{c|}{DDP} & \multicolumn{2}{c|}{DDPEN}\\
           & Forward pass & Backward pass & Forward pass & Backward pass\\
         \hline\hline 
         Free & $\textbf{191}\pm1$ & $\textbf{188}\pm1$ &  $\textbf{191}\pm1$ & $198\pm1$\\  
         Cylinders & \textbf{243}$\pm$1 &$313\pm27$  & 255$\pm8$ &\textbf{281}$\pm32$\\ 
         Cubes without minima & $284\pm54$ &$\textbf{262}\pm31$ & $\textbf{243}\pm15$  & $266\pm6$\\ 
         Cubes with minima & fail & fail & $\textbf{245}\pm7$ & $\textbf{275}\pm20$ \\ 
         \hline
        \end{tabular}
    \end{center}
\end{table*}


%% file: chapters/4_experiments.tex
\subsection{Simulator}
\begin{figure}[hb!]
    \centering
    \includegraphics[width=8.5cm]{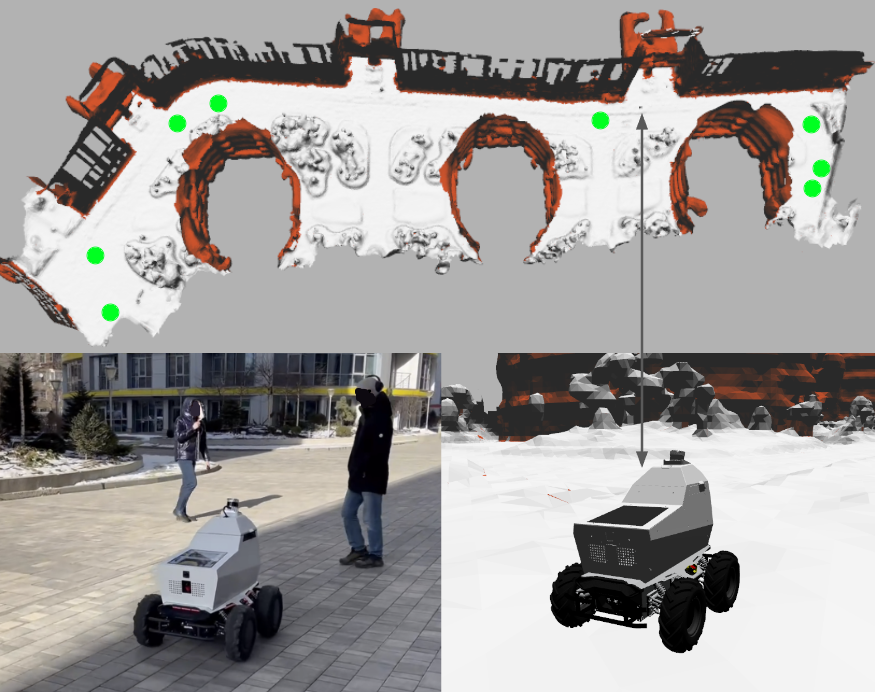}
    \caption{Simulated street on the top on figure, simulated sidewalk robot in environment on the right down and the real robot at the same place on the left down. Green circles on the simulated street show the path points for experiments.} 
    \label{fig::street_and_robot}
\end{figure}

For conducting experiments and evaluation, we use Gazebo [8] simulator with an environment map reconstructed from lidar scans and four wheels sidewalk robot model with a lidar sensor.
Environment map and model of robot is shown in Figure \ref{fig::street_and_robot}.
Point cloud data from simulated lidar sensor after processing by ROS2\cite{ROS2nav2} node provides costmaps that uses as an input to trajectory optimization algorithm and sub goal generation model.
We use a global path passing through 8 points on the reconstructed map for experiments, as shown in Figure. \ref{fig::street_and_robot}. 
$x_{goal}$ is located on the intersection of the local costmap border and the line between next and previous global path points, and is used as input to the trajectory optimization algorithm and sub goal generation model.

\par

\begin{figure*}[ht]
    \centering
    \includegraphics[width=17.7cm]{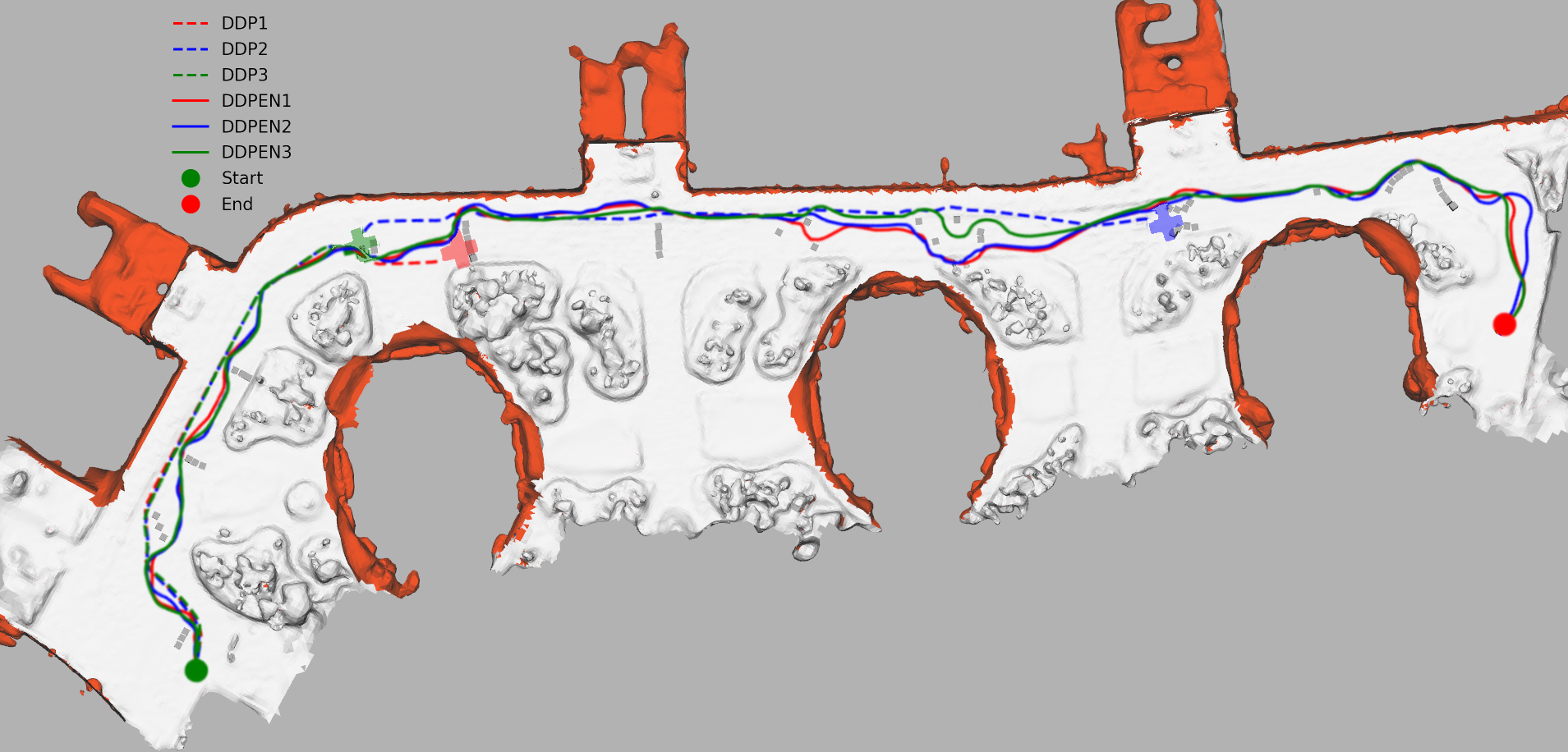}
    \caption{Robot trajectories in a simulated environment with non-convex obstacles. The green circle is the beginning of the paths, the red circle is the end of the paths. The solid and dotted lines are paths created by DDPEN and DDP respectively. The cross marker is the end of the paths in case of hitting the local minimum.} 
    \label{fig::trajectories}
\end{figure*}

\subsection{Running cycle}
\begin{figure}[h]
    \centering
    \includegraphics[width=8.5cm]{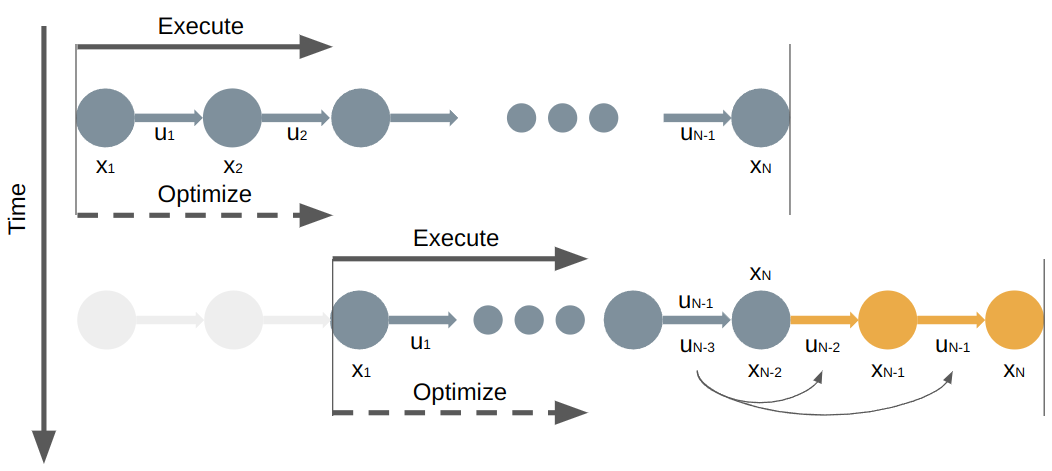}
    \caption{Running cycle - execution trajectory while optimisation process in progress}
    \label{fig::running_cycle}
\end{figure}

The implementation of a particular optimization algorithm takes 75ms for the optimization epoch, where generation of intermediate goal by neural network takes 9.03 ms ± 139 µs.
We use 4 optimisation epochs before executing calculated trajectory. 
For the next cycle of 4 optimisations we use previously generated trajectory with cut past trajectory predictions and added new assumption of trajectory by applying last control, as shown in Figure \ref{fig::running_cycle}.
Executor uses generated trajectory for sending controls to the robot at stable 10Hz and approximate controls between predictions while next optimisation cycle will end.

\par

\section{Evaluation}
We evaluate vanilla DDP and proposed DDPEN trajectory optimization algorithms by executing path of 278 meters in both directions 5 times. 
The path that the robot must follow passes through the green dots as shown in the Fig. \ref{fig::street_and_robot}. 

We manually place different types of obstacles along the path which results in 4 different environments: 
\begin{itemize}
    \item \textit{Free} - original obstacle free environment.
    \item \textit{Cylinders} - environment with additionally placed cylinders with the radius 1m along the path.
    \item \textit{Cubes without minima } - environment with additionally placed cubes with a side 2m forming convex obstacles.
    \item  \textit{Cubes with minima} -  environment with additionally placed cubes with a side 2m forming nonconvex obstacles.
\end{itemize}

Resulting trajectories of DDP and DDPEN  in an environment with \textit{Cubes with minima} is shown in Figure \ref{fig::trajectories}.

Both DDP and DDPEN have equal configuration and dynamic model with linear max velocity limit of 1.5 m/s and angular velocity limit of 0.5 rad/s.
\par
The main goal that we have focused on in this work is the ability to complete the entire path trajectory even with non-convex obstacles, and we also evaluated the path execution time as an indicator of system efficiency.


\par



\par

\subsection{Results}

Table \ref{table:results} shows the average time $\pm$ standard deviation in seconds for passing 4 different scenarios, where  longer execution times mean more iterations were required to converge, failures happens due to weak gradient signals in configurations that are close a local minima.
Table \ref{table:results} shows expected result for DDP which can't complete trajectory on map with local minima, while our proposed method DDPEN is being able to complete trajectory with nonconvex obstacles.

Time of path execution in forward and backward directions are not equal, because trajectories in different directions have minor differences that affect the optimization process.

We can see time execution deviation in the same scenarios that shows how simulation errors affects trajectory optimisation process.
A small error in the the simulation process leads to differences in the execution time of the trajectory, especially in the case of trajectories located near obstacles.

The sum of trajectory execution mean time have differences on cylinders and cubes maps - DDP $1102\pm68$s comparing to DDPEN $1045\pm37$s, because DDPEN starts avoiding obstacles based on additional sub goal provided by Sub goal generation model earlier than obstacle cost $c_{o}$ starts to affect the vanilla DDP optimization process as shown in Figure \ref{fig::DDP_vs_DDPEN}.

\par


\par

%% file: chapters/5_conclusion.tex
\section{Conclusion}
In this paper we have shown a novel approach DDPEN that allows trajectory optimization algorithms to escape from local minima.
Our approach utilizes an additional optimization term representing the direction towards the robot should move in order to escape local minima.
This additional optimization term based at deep sub goal generation model that predicts future directions that the robot should follow in order to avoid local minima.
The model is trained with synthetic dataset and overall system is evaluated on a simulated scene at Gazebo simulator.

The evaluation shows that the proposed method DDPEN is more reliable against the local minima problem, while vanilla DDP fails.

